\documentclass[]{fairmeta}
\usepackage{makecell}
\usepackage{wrapfig}
\usepackage{tabularx}
\usepackage{textcomp}
\usepackage{stfloats}
\usepackage{url}
\usepackage{verbatim}
\usepackage{titlesec}
\usepackage{tocloft}
\usepackage{adjustbox}
\usepackage{multirow}
\usepackage{pifont}
\usepackage[sc]{mathpazo}
\usepackage{tikz}
\usepackage{comment}
\usepackage{amsmath,amssymb}
\usepackage{colortbl}
\usepackage{natbib}
\usepackage{color}
\usepackage{booktabs} 
\usepackage{hyperref}
\usepackage{graphicx}
\usepackage{subcaption}
\usepackage{multirow}
\usepackage{subcaption}
\RequirePackage{xspace}
\makeatletter
\DeclareRobustCommand\onedot{\futurelet\@let@token\@onedot}
\def\@onedot{\ifx\@let@token.\else.\null\fi\xspace}
\usepackage[most]{tcolorbox}
\usepackage{xcolor}
\usepackage{array}
\usepackage{tabularx}
\usepackage{siunitx} 
\usepackage{makecell}
\usepackage[table]{xcolor}
\usepackage{caption}
\definecolor{headerpurple}{HTML}{d8d2fc}
\definecolor{rowgray}{gray}{0.95}

\makeatother

\definecolor{adptorange}{RGB}{248, 205, 172}
\definecolor{cmpblue}{RGB}{189, 215, 238}
\definecolor{cmpblue}{RGB}{189, 215, 238}

\definecolor{our_red}{RGB}{232,157,160}
\definecolor{our_blue}{RGB}{136,206,230}
\definecolor{our_orange}{RGB}{246,200,168}
\definecolor{our_green}{RGB}{178,211,164}

\definecolor{attn_code0}{RGB}{247,215,200}
\definecolor{attn_code1}{RGB}{238,169,139}
\definecolor{mlp_code0}{RGB}{204,201,221}
\definecolor{mlp_code1}{RGB}{102,95,153}
\definecolor{mygray}{HTML}{f0f0f0}

\definecolor{token_blue}{RGB}{84, 120, 140}

\usepackage{pifont}
\usepackage{bbding}
\usepackage{fontawesome}
\usepackage{xspace}

\usepackage{float}

\newlength\savewidth

\newcolumntype{x}[1]{>{\centering\arraybackslash}p{#1pt}}
\newcolumntype{y}[1]{>{\raggedright\arraybackslash}p{#1pt}}
\newcolumntype{z}[1]{>{\raggedleft\arraybackslash}p{#1pt}}

\renewcommand{\paragraph}[1]{\vspace{1mm}\noindent\textbf{#1}}
\usepackage{colortbl}
\usepackage{xcolor}
\usepackage{wrapfig}

\renewcommand{\paragraph}[1]{\vspace{1.25mm}\noindent\textbf{#1}}

\usepackage{algorithm}
\usepackage{listings}

\definecolor{codeblue}{rgb}{0.25, 0.5, 0.5}
\definecolor{codekw}{rgb}{0.35, 0.35, 0.75}
\lstdefinestyle{Pytorch}{
    language = Python,
    backgroundcolor = \color{white},
    basicstyle = \fontsize{9pt}{8pt}\selectfont\ttfamily\bfseries,
    columns = fullflexible,
    aboveskip=1pt,
    belowskip=1pt,
    breaklines = true,
    captionpos = b,
    commentstyle = \color{codeblue},
    keywordstyle = \color{codekw},
}

\definecolor{green}{HTML}{009000}
\definecolor{red}{HTML}{ea4335}

\title{From Pixels to States: Rethinking Interactive World Models as Game Engines}
\author[1,*]{Zhen Li}
\author[1,*]{Zian Meng}
\author[1]{Shuwei Shi}
\author[1]{Mingliang Zhai}
\author[1]{Jiaming Tan}
\author[1,\dagger]{Chuanhao Li}
\author[1,\dagger]{Kaipeng Zhang}

\affiliation[1]{Alaya Lab}

\abstract{
Building interactive worlds that respond coherently to player actions has long been a shared goal of computer graphics, games, and artificial intelligence.
Recent video generative models provide a data-driven route toward this goal by predicting future observations conditioned on user actions, and are increasingly regarded as potential next-generation game engines.
Realizing a genuinely interactive game world, however, requires interaction outcomes that follow rules over evolving game conditions, consequences that persist over long horizons, and a generation loop that operates in real time.
Conventional game engines realize these properties through a recurrent action-state-observation loop, in which player actions update an explicit game state according to predefined rules and observations are rendered from the resulting state.
Taking this loop as an organizing lens, this paper examines interactive game world modeling along four dimensions: player action control, game state dynamics, state-observation persistence, and real-time interactive generation.
For each dimension, we start from the capabilities required by an interactive game world, group existing approaches into representative families, and discuss the strengths and trade-offs of each family.
Complementing this analysis, we present a scalable data engine for \textit{Black Myth: Wukong} that collects over 90 hours of gameplay with frame-aligned player actions, ground-truth game states, and visual observations, together with structured and semantic annotations, as a resource for state-aware game world modeling.
We hope this paper offers a clear picture of where the field stands and fosters progress toward interactive game worlds.
}

\date{\today} 

\begin{document}
\maketitle

\renewcommand{\thefootnote}{\fnsymbol{footnote}}
\footnotetext[1]{Equal contribution.}
\footnotetext[2]{Corresponding authors: chuanhao.li@shanda.com; kaipeng.zhang@shanda.com}


\section{Introduction}

\begin{figure*}[t]
\centering
\includegraphics[width=1.0\linewidth]{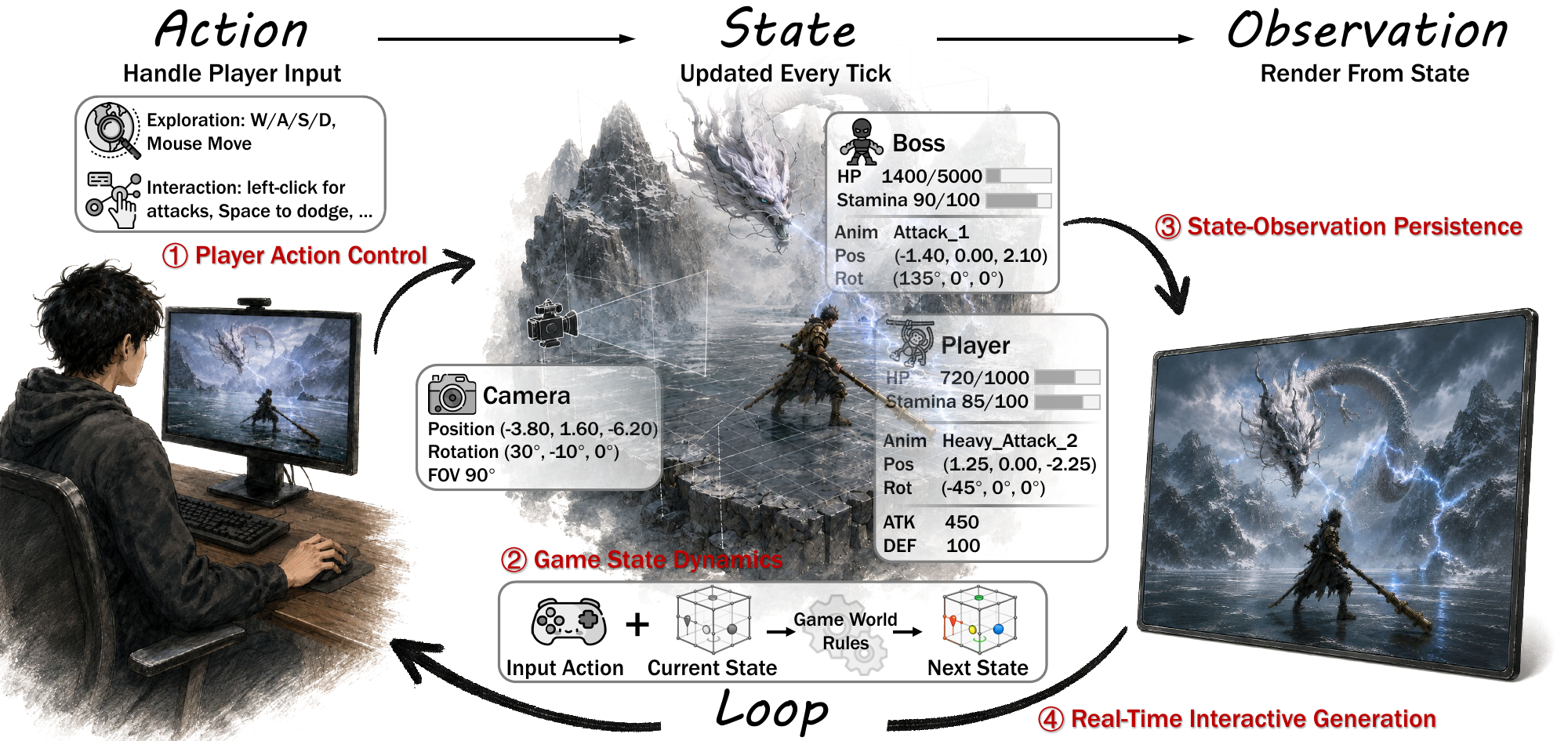}
\caption{
Interactive game worlds as an action-state-observation loop.
Player inputs update the internal game state according to game rules, and the resulting state is rendered into observations that continuously feed back to the player.
}
\label{fig:framework}
\end{figure*}

Building worlds that a person can act within and that respond coherently has long been a shared goal of computer graphics, games, and artificial intelligence.
Recent video generative models have opened a data-driven route to this goal: instead of hand-authoring geometry, assets, and behavior rules, a single model learns the regularities of a world directly from footage~\cite{brooks2024video, zhu2024sora, yue2025simulating} and predicts future observations conditioned on user actions.
Driven by player inputs such as keyboard and mouse, this paradigm has produced increasingly convincing interactive worlds, from real-time neural game engines~\cite{valevski2025diffusion, decart2024oasis, guo2025mineworld} to world simulators built on video generation models~\cite{bruce2024genie, mao2025yume, he2025matrix, parkerholder2025genie}, positioning such models as potential next-generation game engines~\cite{yu2025position}.
These advances have emerged across several related lines of research that have often been pursued separately.
Camera-controlled video generation refines how viewpoint motion is specified and injected~\cite{wang2024motionctrl, he2025cameractrl, li2025hunyuan}; language- and entity-level interfaces broaden the range of interactions a model can express~\cite{che2025gamegen, mao2026yume, zhu2026incantation, xiong2026actworld}; memory mechanisms keep long rollouts consistent with what was generated before~\cite{zhang2025frame, yu2025context, xiao2025worldmem}; and distillation and streaming techniques push generation toward real-time rates~\cite{yin2025slow, liu2026rolling, he2025matrix}.
Together, these lines address key aspects of interactive game world modeling, but their connections and combined implications for game world modeling remain less clear.

In this paper, we organize these methods around the recurrent action-state-observation loop of conventional game engines, as illustrated in Figure~\ref{fig:framework}.
In \textit{Black Myth: Wukong}, for example, when the player presses the attack key, the engine does not directly produce pixels; it first consults the current game state, checking stamina, cooldowns, and whether the boss's position and animation phase leave it vulnerable, then applies the combat rules to commit the outcome, a miss, damage, an interrupted animation, or a phase transition, and only then renders the coming frames from the updated state.
This loop outlines what a generative counterpart must provide, and we accordingly examine interactive game world modeling along four dimensions covering \textbf{player action control}, \textbf{game state dynamics}, \textbf{state-observation persistence}, and \textbf{real-time interactive generation}.
Player action control concerns how player intent is represented before it drives generation, while game state dynamics captures how the underlying condition of the world is represented and evolved under game rules.
The remaining two dimensions address what the loop demands of generation itself: the consequences of interactions must remain consistent over long horizons (state-observation persistence) and must be delivered at the latency of play (real-time interactive generation).
For each dimension, we categorize existing approaches into representative families, summarize what each family offers and at what cost, and analyze how far it carries toward what an interactive game world requires.

Progress along these dimensions also hinges on data, as learning rule-governed state transitions calls for gameplay videos paired with explicit, temporally aligned state annotations, which remain scarce in existing datasets~\cite{che2025gamegen, li2025sekai, yu2025gamefactory, he2025plaicraft, zhou2026omniworld, li2026wildworld}.
To address this, we build a reusable data engine for the AAA action role-playing game \textit{Black Myth: Wukong} and collect boss-encounter gameplay from crowdsourced players with diverse skill levels and play styles.
The resulting dataset contains over 90 hours of gameplay at $1280\times720$ resolution and 30 FPS, with frame-aligned keyboard and mouse inputs, engine-exported game states, RGB frames, and depth maps, and we further textualize the recorded actions and states into slot-structured and semantic captions.

To summarize, our contributions are threefold: 
\begin{itemize}
\item We introduce a unified framework for interactive game world modeling, grounded in the action-state-observation loop of conventional game engines, that organizes the field along four dimensions: player action control, game state dynamics, state-observation persistence, and real-time interactive generation.
\item We systematically review existing approaches along these four dimensions, grouping them into representative families and analyzing the strengths and trade-offs of each family toward interactive game worlds.
\item We build a scalable data engine for \textit{Black Myth: Wukong} and collect over 90 hours of gameplay videos with frame-aligned player actions and ground-truth game states, together with structured and semantic annotations, as a resource for state-aware game world modeling.
\end{itemize}

\section{Related Work}

\subsection{Video Generation Models}

Video generation models have advanced rapidly~\cite{brooks2024video, zheng2024open, wan2025wan, kong2024hunyuanvideo, yang2025cogvideox}, typically coupling a language model that encodes the condition with a diffusion transformer (DiT)~\cite{peebles2023scalable} that synthesizes the video in a latent space~\cite{rombach2022high}.
Existing models fall into three families according to how the information produced by the language model interacts with the DiT.
A first family keeps the language model external and frozen, with its output embeddings read by every DiT block through cross-attention, as in Wan~\cite{wan2025wan} and Open-Sora~\cite{zheng2024open}.
A second family admits the text representations into the DiT's joint attention, realized with modality-specific weights in MM-DiT~\cite{esser2024scaling}, dual streams fused into one in HunyuanVideo~\cite{kong2024hunyuanvideo}, and a concatenated sequence under expert-specific normalization in CogVideoX~\cite{yang2025cogvideox}.
A third family integrates the language model into the backbone as a generative pathway in its own right, whether by training one transformer under both a language loss and a diffusion or flow objective~\cite{zhou2025transfusion, xie2025show, ma2025janusflow}, by decoupling parameters per modality under shared global attention~\cite{liang2025mixture, deng2025emerging}, or, at the extreme, by absorbing generation entirely into next-token prediction~\cite{wang2024emu}.
These models provide the generative backbones upon which the interactive game world models discussed below are built.

\subsection{Interactive Game World Models}

Game world models take player inputs, drive the underlying evolution of the world, and deliver the outcome as visual observations~\cite{yu2025survey, liu2026towards}, and are regarded as potential next-generation game engines~\cite{yu2025position}.
Early systems train bespoke generators on a single game, from action-conditional prediction in Atari~\cite{oh2015action}, GameGAN~\cite{kim2020learning}, and world models for agent training~\cite{ha2018world, alonso2024diffusion, hafner2025training} to neural game engines such as GameNGen~\cite{valevski2025diffusion}, Oasis~\cite{decart2024oasis}, MineWorld~\cite{guo2025mineworld}, and WHAM~\cite{kanervisto2025world}.
Video generation models then make generated worlds explorable, from latent-action models~\cite{menapace2021playable, bruce2024genie} and Yume~\cite{mao2025yume} to engines pushing generation toward real-time rates, such as The Matrix~\cite{feng2025matrix}, Hunyuan-GameCraft~\cite{li2025hunyuan}, Matrix-Game~2.0~\cite{he2025matrix}, and Genie~3~\cite{parkerholder2025genie}, though interaction remains largely confined to viewpoint and locomotion.
Recent systems restore game interactions through language-specified events and instructions~\cite{che2025gamegen, tang2025hunyuan, mao2026yume, team2026dreamx, parkerholder2025genie}, scene-generalizable action control~\cite{yu2025gamefactory}, spatially localized device controls~\cite{tong2026scope}, and object- and entity-level interfaces~\cite{xiong2026actworld, zhu2026incantation, wang2026reactivegwm}.
In parallel, general-purpose world models extend to the game domain~\cite{xiang2025pan, zhu2026astra}, game-trained control transfers to real-world footage~\cite{sun2025virtual}, and systems efforts push the loop toward real time~\cite{wang2026matrix, zhao2026minwm, team2026inspatio}.
An emerging line further makes the game state explicit, through per-frame state annotations~\cite{li2026wildworld}, stepwise state prediction~\cite{cheng2025animegamer}, persistent 3D state~\cite{garcin2026beyond}, and executable symbolic rules~\cite{zhao2026neuro}.
Together, these systems trace a trajectory from single-game simulators toward general, interactive game worlds.
The remainder of this paper revisits these methods along the four dimensions introduced above, organizing each dimension into representative families and discussing their strengths and trade-offs.

\section{Interactive Game World Modeling as Game Engine}

Interactive game world modeling concerns worlds in which future observations are driven by player inputs, mediated by evolving game conditions, and delivered as real-time visual feedback.
Following the action-state-observation loop of conventional game engines, we structure this problem along four dimensions: player action control, game state dynamics, state-observation persistence, and real-time interactive generation.
Each subsection first states what an interactive game world requires on one dimension, then groups representative approaches into families and discusses the strengths and trade-offs of each family.

\subsection{Player Action Control}

Player action control defines how player intent is represented before it drives generation.
In action role-playing games, player inputs are heterogeneous: some control navigation and viewpoint, while others trigger semantic game events such as attacks, dodges, skills, or item usage.
We accordingly group existing methods by whether intent is expressed as geometric trajectories, motor signals, or semantic events.

\paragraph{Actions as geometric trajectories.}
This category represents player intent as explicit geometric quantities, typically continuous pose trajectories, known as camera-controlled video generation.
One group treats camera motion as an external conditioning signal, encoded as rotations and translations in MotionCtrl~\cite{wang2024motionctrl} or as Plücker ray embeddings in CameraCtrl~\cite{he2025cameractrl}, with CamCo~\cite{xu2024camco} and CamI2V~\cite{zheng2024cami} extending such control to image-to-video generation under epipolar-constrained attention.
The other builds camera geometry into the architecture as a prior, modeling relative frustum geometry~\cite{li2025cameras}, unifying positional encoding over poses, intrinsics, and distortion~\cite{zhang2026unified}, or introducing viewing-ray rotary encoding for long-horizon view consistency~\cite{xiang2026geometry}.
These methods deliver precise and consistent viewpoint control, yet they cover only navigation rather than broader player interaction, and the precise trajectories they require remain distant from natural user inputs.

\paragraph{Actions as motor signals.}
Motor signals encode player intent as the operations themselves, with an action space either defined by the control device or learned from data.
For device-defined actions, methods differ mainly in the injection mechanism: Oasis~\cite{decart2024oasis} and the Matrix-Game series~\cite{zhang2025matrix, he2025matrix} embed frame-wise keyboard and mouse states as conditioning vectors, MineWorld~\cite{guo2025mineworld} and iVideoGPT~\cite{wu2024ivideogpt} interleave actions with visual tokens as action tokens or action-bearing slots, Hunyuan-GameCraft~\cite{li2025hunyuan} maps discrete key states into a continuous camera representation, and SCOPE~\cite{tong2026scope} confines localized events such as firing to their affected regions.
In the absence of action annotations, latent-action models instead learn the action space directly from unlabeled videos~\cite{menapace2021playable, menapace2022playable, bruce2024genie, garrido2026learning, zhang2026dila, jiang2026olaf, qiu2026self}.
These representations stay faithful to how players natively operate, and atomic controls such as moving or firing can be learned reliably from large-scale supervision.
Nonetheless, raw signals underdetermine intent, as the same input can express different intents depending on the game state, and such ambiguity makes composite operations, as in combo skills whose meaning exceeds the individual key presses, harder to learn.

\paragraph{Actions as semantic events.}
Semantic events state player intent explicitly in natural language, with methods differing mainly in the granularity at which language is grounded.
Scene-level control extends the prompt-to-content paradigm into the interactive loop, steering rollouts with textual instructions or promptable world events~\cite{che2025gamegen, parkerholder2025genie, mao2026yume, tang2025hunyuan}, and Pandora~\cite{xiang2024pandora} and PAN~\cite{xiang2025pan} further adopt stepwise free-text actions as the native control channel of general world models.
Finer-grained methods ground language over individual entities or specific moments of the rollout: Promptable Game Models~\cite{menapace2024promptable} pioneer per-agent control with natural-language actions and goals, Incantation~\cite{zhu2026incantation} controls multiple entities simultaneously through per-chunk, per-entity conditioning, ReactiveGWM~\cite{wang2026reactivegwm} steers high-level NPC strategies decoupled from player control, and ActWorld~\cite{xiong2026actworld} extends navigation-centric models to object-level interaction via action-aware memory.
These representations make intent explicit and extend control to interactions that no fixed input scheme covers, yet the conditioning cost grows with grounding granularity, imposing a trade-off between control fidelity and interactive efficiency.

\subsection{Game State Dynamics}

In game engines, the game state is a set of explicit variables, such as health, stamina, cooldowns, and equipment, that describes the current status of the world, and the engine continually updates it from the player action according to game rules.
Game state dynamics thus concerns how a model represents this state and realizes its update, and existing methods differ in the form in which the state exists: entangled in observations, compressed into learned latents, or maintained as explicit descriptions~\cite{wang2026mechanistic}.

\paragraph{States entangled in observations.}
The prevailing design maintains no separate state and lets world dynamics live entirely in pixels.
Modern video foundation models such as Open-Sora~\cite{zheng2024open}, Wan~\cite{wan2025wan} and LTX-2~\cite{hacohen2026ltx} internalize physics and object interactions from internet-scale video and are increasingly regarded as implicit world simulators~\cite{zhu2024sora, yue2025simulating, wiedemer2025video}.
Game world models inherit this state-free formulation, predicting future frames directly from recent observations and actions, from early frame prediction in Atari~\cite{oh2015action, chiappa2017recurrent} through GameNGen~\cite{valevski2025diffusion}, DIAMOND~\cite{alonso2024diffusion}, and WHAM~\cite{kanervisto2025world} to open-ended interactive models driven by native keyboard and mouse inputs~\cite{decart2024oasis, zhang2025matrix, he2025matrix}.
This design inherits the full visual capacity of video foundation models, yet the state exists only as a recent-observation window, so world rules are captured as pixel correlations, making spatial and logical consistency hard to guarantee~\cite{kang2025how}.

\paragraph{States as learned latents.}
A second design maintains a compact latent state that evolves recurrently under actions.
This formulation originates from world models for reinforcement learning~\cite{ha2018world}, and games have been its proving ground, from GameGAN~\cite{kim2020learning} simulating Pac-Man with a memory-equipped latent dynamics engine to the Dreamer lineage~\cite{hafner2019learning, hafner2025mastering, hafner2025training} training agents inside latent dynamics models, from continuous control to Atari and Minecraft.
The same design prevails in embodied planning~\cite{zhou2025dino, bar2025navigation}, and recent video world models carry a state-space summary of the entire interaction history into diffusion for long-context rollouts~\cite{savov2025statespacediffuser, yu2025videossm}.
Latent states require no in-game annotation and thus scale readily with in-the-wild video, but they lack interpretability and, being learned from visual prediction, remain unreliable for visual changes driven by non-visual causes, as when the same strike kills or merely wounds depending on the remaining health of the target.

\paragraph{States as explicit descriptions.}
A third formulation writes the state out in symbolic or textual form, turning state transition into reasoning.
Large language models have been used as textual world simulators that predict the next state for planning agents~\cite{hao2023reasoning, wang2024can}, and multimodal variants extend such transitions across modalities~\cite{ge2024worldgpt, xiang2025pan}.
In games, however, explicit states have so far served mainly as data and evaluation, with WildWorld~\cite{li2026wildworld} providing per-frame state annotations extracted from engine memory and EgoCS-400K~\cite{guo2026egocs} aligning gameplay videos with player states and events; the closest step, AnimeGamer~\cite{cheng2025animegamer}, predicts character attributes such as stamina at every step, though this state does not directly drive the generation.
Explicit states are readable and verifiable, and their transitions align with what language models excel at.
On the other hand, maintaining such states presupposes large-scale state-annotated gameplay data, which remain scarce, and discrete records capture the continuous visual dynamics of a game only partially, leaving their integration into the generation loop largely unexplored.

\subsection{State-Observation Persistence}

State-observation persistence requires generated observations to remain consistent with the evolving game state over long horizons.
Video-based world models retain nothing beyond the recent context, so a scene may silently change once it leaves view~\cite{ma2026out, ye2026mind, zhang2026mbench}; games demand more, since transitions can be irreversible and the state keeps evolving off screen, as when a stray strike leaves lasting damage, or the boss returns to view in its second-phase form.
Long-horizon memory mechanisms are introduced to carry history into generation, and we group them according to whether memory records the past or estimates the present.

\paragraph{Memory as stored observations.}
The prevailing family carries stored observations and differs only in the retrieval key.
Temporally indexed methods keep history as a sequence, packing long histories with progressively stronger compression of older frames~\cite{zhang2025frame}, mapping them to short embeddings~\cite{zhang2025tinyhistory} or pose-free fixed-budget memories~\cite{wu2026infinite}, folding the sequence into a running summary~\cite{po2025long, yu2025videossm, chen2025recurrent}, or learning the retrieval itself~\cite{cai2026mixture, yu2026memlearner}.
Spatially indexed methods recall history by where it was observed, selecting context frames by field-of-view overlap~\cite{yu2025context}, indexing memory banks by camera poses and timestamps~\cite{xiao2025worldmem, oshima2025worldpack}, encoding absolute poses into compressed attention caches~\cite{hong2025relic}, or anchoring past views to surfels~\cite{li2025vmem}.
These approaches are simple and dependable, grounding revisited views in real evidence and performing well in largely static scenes.
In interactive game worlds, however, the world changes continually, so a faithful copy of the past may not serve as the reference for the present: although Spatia~\cite{zhao2026spatia} excludes dynamic entities from memory, the environment itself can also change, as when a skill demolishes a nearby building, yet on revisit the memory may restore the intact one.

\paragraph{Memory as estimates of the present.}
A newer family starts from the fact that the world keeps changing while out of view, and therefore updates memory dynamically.
ActWorld~\cite{xiong2026actworld} retains the frames recording event transitions that determine subsequent object states, Hybrid Memory~\cite{chen2026out} tracks dynamic subjects through out-of-view intervals, flow-equivariant models~\cite{lillemark2026flow} propagate the dynamics of unobserved regions, PERSIST~\cite{garcin2026beyond} advances an entire latent 3D state under scene dynamics and generates guidance frames from it, and AnimeGamer~\cite{cheng2025animegamer} predicts an explicit game state at every step, though the state does not directly condition the generation.
These methods leave memory updates to the adaptive judgment of the learned model, proving effective in their respective domains.
In game world modeling, however, high-frequency interactions induce small changes that accumulate into substantial divergence, demanding update decisions that are efficient and reliable at every step, a requirement for which a principled grounding remains an open question.

\subsection{Real-Time Interactive Generation}

Real-time interactive generation requires player inputs to be turned into observations at interactive rates, and two latencies with opposite requirements bound this loop.
Control latency, from an input to its visible response such as the viewpoint turning, should be minimized, as it governs game feel.
Consequence latency, from an action to its rule-governed outcome such as a strike landing, should instead be accurate, since an attack unfolds through startup, active, and recovery, and an outcome surfacing instantly reads as wrong.
Existing works treat every delay as a quantity to shrink, and we review them according to the delay each reduces, the time to produce the next frame or the time for a changed condition to take effect.

\paragraph{Reducing generation latency.}
A first family shortens the time from an input to the next frame.
Step-reduction methods distill a multi-step bidirectional teacher into a few-step causal student, a recipe established by CausVid~\cite{yin2025slow} with asymmetric distribution matching distillation and refined through causal-teacher initialization~\cite{zhao2026causal}, streaming-aware distillation~\cite{liu2026streaming}, and subsequent reinforcement learning~\cite{zhang2026astrolabe}.
Streaming methods restructure generation into a causal stream, jointly denoising a rolling window at increasing noise levels~\cite{liu2026rolling}, sharing a noise level across blocks for pipelined inference~\cite{zou2026hiar}, simulating inference-time history errors during training~\cite{guo2026end}, or combining action injection with few-step distillation to stream at real-time rates~\cite{he2025matrix}.
Systems efforts reduce wall-clock cost without changing the sampler, through full-stack training and inference pipelines~\cite{zhao2026minwm}, independent frame generation~\cite{team2026inspatio}, and hardware-algorithm co-design~\cite{zeng2026scalable}.
These mechanisms have brought high-quality generation within reach of interactive frame rates, yet they uniformly shorten the path from input to frame, and none concerns when an outcome should appear.

\paragraph{Reducing conditioning latency.}
A second family lets the conditioning change during a rollout without regenerating the stream.
Yume-1.5~\cite{mao2026yume} switches text-controlled world events under streaming acceleration, MagicWorld~\cite{li2025magicworld} evolves scenes under keyboard-driven navigation, and DreamX-World~\cite{team2026dreamx} tunes composable event instructions for general interactive control.
These systems show that responsiveness is not sampling speed alone, since the conditioning itself must track user intent.
The timing of a triggered outcome, however, is left to the generative prior, with no mechanism aligning it with the rule- or animation-defined moment of the action.
\section{Data Engine}

\begin{figure*}[t]
\centering
\includegraphics[width=0.9\linewidth]{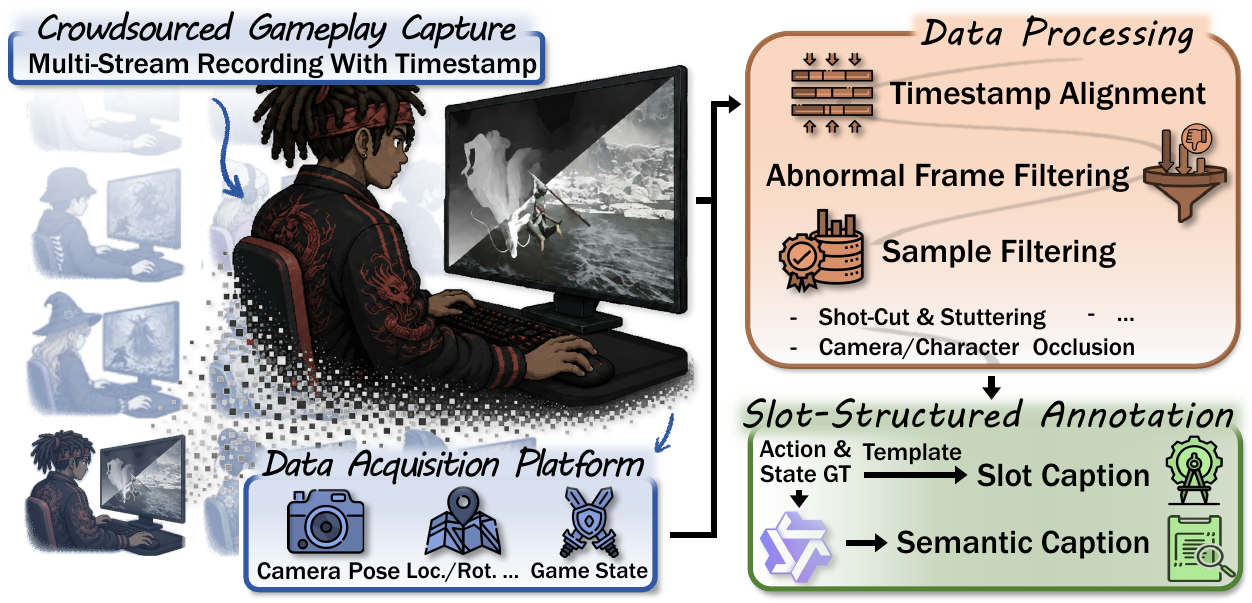}
\caption{The dataset curation pipeline.}
\label{fig:data-framework}
\end{figure*}

Progress along the four dimensions above, in particular toward explicit game state dynamics, requires gameplay videos paired with explicit state annotations, yet such data remain scarce in existing datasets~\cite{che2025gamegen, li2025sekai, yu2025gamefactory, he2025plaicraft, zhou2026omniworld}.
To address this, we build a data engine for \textit{Black Myth: Wukong}, a photorealistic action role-playing game built on Unreal Engine.
Specifically, we focus on boss encounters, which feature frequent interactions such as attacks and dodges, as well as continuous state transitions.
Each encounter also lasts several minutes, providing long rollouts for training and evaluating long-horizon generation.
The character is controlled by crowdsourced players with diverse skill levels and play styles, producing varied behaviors through the game's native keyboard and mouse interface.

The resulting dataset contains over 90 hours of gameplay at $1280\times720$ resolution and 30 FPS along with three types of frame-aligned ground truth: actions captured as raw mouse and keyboard inputs from the player, states describing the underlying evolution of the game world, and observations comprising the RGB frames and depth maps.
Concretely, the states cover the poses of the camera, the player character, and the boss, as well as the characters' current animations, active skills, and gameplay attributes (health, stamina, attack/defense, equipment, etc.).
We organize the data pipeline into two stages, \textbf{Crowdsourced Gameplay Capture} and \textbf{Data Processing and Slot-Structured Annotation}, where the capture stage introduces minor overhead on the crowdsourcing side and requires no additional hardware, enabling scalable data collection at controlled cost.

\subsection{Crowdsourced Gameplay Capture}
We instrument the game engine to export interaction data at every tick as a structured stream, where each record covers the actions and states, serialized in JSON format and written to a local file.
Every frame thus carries its complete interaction context.
For observations, we develop a split-screen recording scheme based on ReShade and OBS Studio.
A custom ReShade shader partitions the display into sub-windows that present the RGB frames and depth maps from the rendering buffer.
The in-game UI is removed by disabling the corresponding rendering passes, yielding clean frames of the game world.
We further adapt OBS Studio to log the system timestamp of each recorded frame, which, together with the engine-side records stamped by the same clock, provides a common temporal basis for subsequent frame-level alignment.
To standardize capture across heterogeneous crowdsourcing machines, we design a tool suite that automatically enforces unified in-game settings and recording configurations.

\subsection{Data Processing and Slot-Structured Annotation}
The processing stage consolidates the recorded streams into training-ready samples.
The streams are first aligned frame by frame through the recorded system timestamps, segmented into per-encounter samples, and re-encoded into training-friendly formats.
Samples exhibiting dropped frames, stuttering, or cross-stream inconsistency are discarded in their entirety.
Notably, the game engine encodes actions and states numerically, as internal identifiers (\emph{e.g.}, action IDs) and absolute quantities (\emph{e.g.}, world coordinates), which are precise but semantically opaque.
We therefore annotate each sample with two forms of captions that retain the precision and supply the semantics, respectively.
The first is the slot caption, which aggregates the per-tick records within fixed-length time windows and textualizes each window into structured action and state slots.
The second is the semantic caption, generated by Qwen3-VL-235B-A22B-Instruct from sampled frames together with the corresponding action and state records as context, describing the executed actions and the resulting state transitions.
The resulting captions can serve as next-state supervision for language-based state transition and as state conditioning for generation, supporting future state-aware game world models.
\section{Conclusion}

In this paper, we examined interactive game world modeling through the recurrent action-state-observation loop of conventional game engines, organizing existing approaches into representative families along four dimensions: player action control, game state dynamics, state-observation persistence, and real-time interactive generation.
The analysis shows that existing methods already deliver natural input control, explorable scenes, and generation approaching real-time rates, whereas the capabilities that remain difficult, determining outcomes from accumulated game conditions, preserving consequences beyond the current view, and surfacing effects at rule-defined moments, all revolve around the game state, which most models keep implicit.
We further presented a scalable data engine for \textit{Black Myth: Wukong}, contributing over 90 hours of gameplay with frame-aligned actions, engine-exported states, and visual observations to ease the data scarcity that limits state-aware approaches.
Looking forward, integrating explicit game state into the generation loop, grounding memory updates in state transitions, and aligning outcome timing with game rules remain open challenges, and we hope this paper charts these directions and fosters progress toward interactive game worlds.

\bibliographystyle{abbrv}
\bibliography{references}

\end{document}